\documentclass[runningheads]{llncs}
\usepackage[T1]{fontenc}
\usepackage{graphicx} 
\usepackage{array}
\usepackage{subfigure} 
\usepackage{booktabs}
\usepackage{epstopdf}
\usepackage[export]{adjustbox} 
\usepackage{amsmath}
\usepackage{enumitem}
\usepackage[ruled,vlined]{algorithm2e}
\usepackage{url} 
\usepackage{float}
\usepackage{hyperref} 
\begin{document}
\title{Depth-Aware Path Planning and Collision Avoidance for Robotic Nursing Assistants Using Real-Time 3D Vision \thanks{This work was supported by the Technology Innovation Hub (TIH) for Cobotics, Indian Institute of Technology Delhi, under the project titled “Healthcare Robotics,” reference number IITM/IHFC-IIT DELHI/LB/370.}}
\titlerunning{Depth-Aware Path Planning for Robotic Nursing Assistants}
%
\author{Peeyush Thakur\inst{1}\orcidID{0009-0002-5432-9699} \and
Dharmendra Sharma\inst{1}\orcidID{0009-0009-1361-1189} \and
Archit Sharma\inst{1}\orcidID{0009-0005-8926-2086} \and Sandeep Gupta\inst{2}\orcidID{0000-0001-9876-8859} \and Anuj Nandanwar\inst{1}\orcidID{0000-0002-6094-0905} \and Narendra Kumar Dhar\inst{1}\orcidID{0000-0002-5343-5812} \and Laxmidhar Behera\inst{1,2}\orcidID{0000-0003-4102-3834} }
\authorrunning{P. Thakur et al.}
%

\institute{Indian Institute of Technology Mandi, India\\
\email{s23070@students.iitmandi.ac.in}
\and
Indian Institute of Technology Kanpur, India}

\maketitle              
\begin{abstract}
This paper presents a depth-aware path planning and collision avoidance framework for the safe navigation of robotic nursing assistants in cluttered environments with low-height obstacles. Conventional LiDAR-based navigation systems, while accurate for mapping and localization, fail to detect transparent or low-height obstacles, whereas vision-based systems are prone to noise and occlusion. The proposed framework integrates RGB-D depth sensing, rule-based obstacle avoidance, and the NavfnROS global path planner to drive the robot from an initial point to a goal point. The proposed approach is implemented on the Husky A200 robot, equipped with a Kinova Gen3 arm, to form a complete setup for a robotic nursing assistant. This system features a multi-layer architecture for global planning, perception fusion, and real-time control in the ROS-Gazebo environment. Experimental evaluations in a simulated cluttered environment demonstrate up to 40\% higher success rate and 35\% improvement in minimum clearance compared with single-sensor baselines. Furthermore, real-world experimental results validate the proposed approach as a computationally efficient and safety-aware navigation solution for autonomous healthcare robots. The code is available at \url{https://github.com/Peeyush331098/Depth-based-obstacle-avoidance.git}

\end{abstract}
\section{Introduction}
The rapid advancement of service robotics has opened new possibilities for robotic nursing assistants to support healthcare professionals in performing routine tasks such as medication delivery, patient monitoring, and logistics management. Ensuring autonomous and safe navigation in hospital environments, however, remains a formidable challenge. Unlike structured industrial settings, hospitals are unstructured, dynamic, and highly cluttered, with narrow corridors, movable beds, trolleys, and the constant presence of humans. Under such conditions, reliable path planning and obstacle avoidance are essential to prevent collisions, maintain operational efficiency, and ensure patient safety. Consequently, intelligent perception and navigation algorithms have become the cornerstone of modern healthcare robotics, enabling robots to adaptively plan trajectories, localize themselves, and interact safely with human co-workers in real time.

Among the sensing modalities investigated for autonomous navigation, systems based on LiDAR are the state of the art because of their capacity to provide reliable 3D mapping, high-precision range measurements, and steady performance across a range of lighting conditions. Because of these benefits, LiDAR is an essential part of autonomous wheeled systems and hospital-grade SLAM frameworks \cite{Desouza_2,Fan_5}.  For example, hospital service robots have effectively used 2D LiDAR navigation frameworks for waypoint-based path tracking and corridor following, while 3D LiDAR extensions improve spatial awareness in semi-structured interior spaces \cite{gul_1}.  However, safety in patient-care scenarios may be jeopardized because LiDAR-only systems frequently miss challenging features such as translucent partitions, low-height obstacles, or partially occluded objects \cite{Wang_4}. LiDAR technology also has inherent limitations, such as a limited vertical field of view, low point density, absence of semantic or color information, and difficulty in recognizing translucent or reflective surfaces (such as glass doors or polished floors) \cite{nenebi2025lidar}. These limitations hamper reliable perception in reflective or visually complex healthcare environments.

To overcome such limitations, depth cameras have emerged as a powerful complementary sensing approach \cite{singh2020multivariable}. By providing dense pixel-wise depth maps integrated with visual cues, depth cameras enable enhanced perception of low-lying, irregular, and specular obstacles often missed by LiDAR alone. Therefore, depth imaging constitutes a promising pathway toward perception-aware and context-adaptive navigation, facilitating safer and more efficient path planning \cite{nandanwar2020fuzzy} for robotic nursing assistants in cluttered and dynamic hospital environments. Recent studies demonstrate that exploiting the contextual richness of depth imagery significantly enhances both obstacle perception and path replanning under dynamic conditions \cite{Wang_4,Fan_5}. Moreover, deep reinforcement learning (DRL) and convolutional neural networks (CNNs) have been applied to interpret fused sensor data for adaptive, human-aware navigation \cite{gul_1,Kolar_3}. Vision-based obstacle avoidance algorithms have been proposed in \cite{acquaah2025integrating,machkour2023monocular,adachi2019visual,zhang2021rgb}.

In \cite{acquaah2025integrating}, the authors used YOLOv5 for real-time object detection and distance estimation, which is valid only for known obstacles. In real-world scenarios, there are obstacles that are not recognized by such object detection algorithms. Despite these advances, achieving real-time fusion, maintaining computational efficiency, and ensuring safe human–robot interaction remain open challenges. These issues continue to motivate the development of hybrid perception frameworks that exploit the complementary strengths of LiDAR and depth cameras for robust navigation in healthcare settings \cite{Desouza_2,Kolar_3,Wang_4,Fan_5}.
Hence, there exists a critical research gap in developing a computationally efficient, real-time hybrid perception and path-planning framework that integrates LiDAR and depth camera sensing for context-aware, safe, and adaptive navigation of robotic nursing assistants in cluttered and dynamic hospital environments. Addressing this gap would significantly advance the reliability and autonomy of service robots in healthcare applications.

This paper presents a hybrid depth-aware navigation framework that uses Navfn global planning and a five-zone depth-based reflex layer to avoid obstacles in real time. The framework automatically detects when the Husky A200 robot is stuck and replans a new route, ensuring safe, collision-free motion in cluttered indoor spaces. Experimental results indicate substantial enhancements in navigation success rates and clearance when contrasted with single-sensor methodologies.


\section{Related Work}
Conventional mobile robot navigation frameworks, such as the ROS Navigation Stack, use a global planner (e.g., NavfnROS) to generate paths and a LiDAR-driven local planner (e.g., DWAPlannerROS) to avoid local obstacles \cite{zheng2021ros}.
 LiDAR sensors work well in controlled settings, but they frequently miss low-height or transparent obstacles, making them risky to use in areas where people are present.
 Recent studies have investigated vision-based obstacle avoidance \cite{wenzel2021vision} and RGB-D camera-based navigation \cite{henry2012rgbd}, which offer richer spatial information but suffer from occlusion and depth noise.
 For dynamic indoor situations, hybrid approaches that combine LiDAR and RGB-D perception with global planning have demonstrated promise \cite{altawil2025mobile}.
 The majority of current research, however, focuses on neural models or computationally demanding sensor fusion that is difficult to run in real time.
 In contrast, the proposed Navfn-Reflex framework balances perception accuracy, computing efficiency, and safety for healthcare-oriented mobile robots by achieving quick, rule-based obstacle avoidance through the use of a five-zone depth partitioning and automatic replanning.

\section{Proposed Approach}
The proposed depth-based navigation methodology integrates multiple sensing and planning components to achieve robust, safe, and adaptive navigation for robots in dynamic indoor environments, such as hospitals. The framework uses RGB-D (depth) sensing with a global path planner, enabling the system to perceive complex surroundings, avoid obstacles, and dynamically replan when necessary. 

The methodology consists of the following interconnected stages. First, a map generation process constructs an occupancy grid of the environment using LiDAR-based SLAM. This map forms the foundation for path planning, which combines a global planner for long-term route generation with a local layer for real-time obstacle handling. Sensor processing describes how depth camera data are analyzed and combined into a unified risk signal, and the design choices that make this processing robust are then outlined. The depth-based navigation algorithm integrates all modules into a single operational procedure, and finally the overall workflow describes the hierarchical interaction among the global planning, perception, and local control layers.

Together, these components form a depth-based navigation framework that ensures the robot can navigate autonomously and safely in complex environments populated by humans with dynamic obstacles.

\subsection{Map generation}
\begin{figure}[!ht]
    \centering
    \subfigure[]{\includegraphics[width=0.44\textwidth]{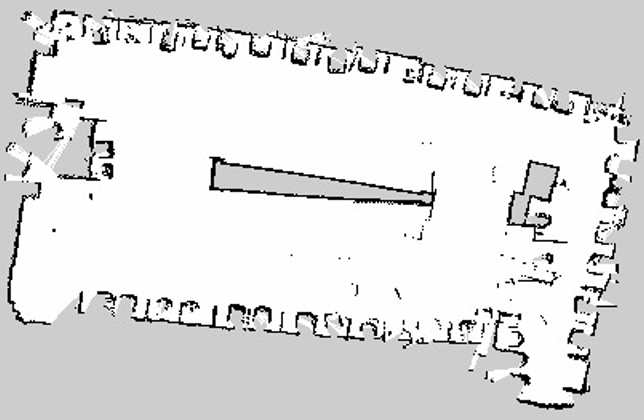}}
    \hfill
    \subfigure[]{\includegraphics[width=0.49\textwidth]{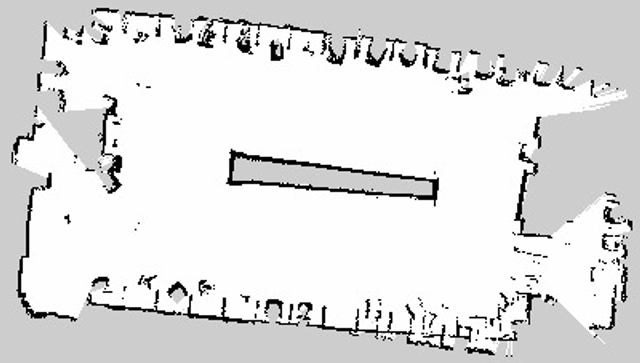}}
    \vspace{-0.25cm}
    \caption{(a) Raw occupancy map and (b) Filtered occupancy map.}
    \label{fig:two_figs}
\end{figure}

The Husky A200 mobile robot constructs a two-dimensional occupancy grid map using a UST-10 LiDAR sensor. LiDAR scans are fused with odometry and IMU data through a SLAM framework. Recent work proposes jointly optimizing robot poses and occupancy map values in one formulation, yielding higher map fidelity and trajectory accuracy~\cite{wang2025occupancy}. As the robot moves, each new scan is aligned to previous data to incrementally build a consistent global map of free, occupied, and unknown regions. The resulting occupancy grid is saved using the \texttt{map\_saver} utility and subsequently used by AMCL for pose estimation during autonomous navigation. The generated raw and filtered maps from the laboratory environment are shown in Fig.~\ref{fig:two_figs}.
\vspace{-.25cm}
 \vspace{-0.5cm}
\subsection{Path planning}
Using the generated occupancy map, the navigation framework adopts a hybrid planning architecture integrating global and local layers. The global planner (e.g., \texttt{NavfnROS}) computes an optimal route over the occupancy grid, while the local planner uses real-time sensor (RGB-D/LiDAR) feedback to handle dynamic obstacles. A corridor-detection mechanism enables safe traversal in narrow passages, and when the global path is obstructed, the system triggers local avoidance and dynamic replanning. Hybrid planning strategies combining global and local optimization have been shown in recent studies to significantly improve trajectory smoothness, collision rates, and responsiveness in changing environments~\cite{yuan2025bio}. This depth-aware, adaptive planning framework enables safe, efficient, and human-aware navigation in complex indoor corridors.

\subsection{Sensor processing}
In \textit{depth camera processing}, a
bottom region of interest (ROI), covering the lower $\rho$ fraction of the depth frame, focuses on obstacles along the robot’s path. The ROI is divided into five zones (\textit{far-left} to \textit{far-right}), ignoring invalid or floor-level pixels. Each zone's mean and 5th-percentile depths yield a stable proximity measure, and the minimum distance $d_{\min}$ over all zones indicates the nearest obstacle. If $d_{\min} < \delta_s$, where $\delta_s$ is the safe distance, a reflex controller issues one of three avoidance actions---\textit{turn left}, \textit{turn right}, or \textit{stop}---temporarily overriding the nominal path follower. \par
In the \textit{fusion and decision layer}, the navigation node combines global planning (\textit{NavfnROS}) with local reflex control through a weighted risk model:
\begin{equation}
R = w_G \sigma_G + w_R \sigma_R,
\end{equation}
where $\sigma_G$ and $\sigma_R$ denote global path deviation and local proximity risk, respectively, and $w_G + w_R = 1$. When $R > R_{\text{th}}$, where $R_{\text{th}}$ is the risk threshold, the system executes a short reflex action, pauses forward motion, and republishes the last goal for automatic replanning. This unified strategy ensures smooth, collision-free navigation and quick recovery in dynamic or cluttered environments.

\subsection{Design choices and robustness}
To ensure dependable performance in the presence of sensor uncertainty, a number of design improvements were integrated into the depth-based perception framework.
 (a) \textit{Percentile-based depth estimation} provides consistent range measurements by reducing noise from hot pixels, specular reflections, and surface discontinuities.  (b) For efficient obstacle detection in confined spaces, \textit{adaptive spatial weighting} dynamically adjusts the contribution of image regions, giving near-field areas more weight. (c) To ensure temporal consistency across the perception pipeline, \textit{temporal validation and time gating} reject outdated or asynchronous frames using timestamp verification. (d) \textit{Temporal fusion with short-term prediction} smooths successive depth frames and maintains spatial continuity during short sensor dropouts.

 Together, these mechanisms provide a robust, uncertainty-tolerant depth perception architecture that enables consistent and reliable navigation in complex indoor environments.

\subsection{Depth-based navigation algorithm}
The complete navigation process is summarized in Algorithm~\ref{alg:navfn_reflex}. The algorithm integrates sensor fusion, path following, obstacle avoidance, and replanning into a single operational flow.
\begin{algorithm}[!ht]
\caption{Depth-based Navfn path follower with reflex safety layer}
\label{alg:navfn_reflex}
\KwIn{Map $\mathcal{M}$, goal $\mathbf{G}=(x_g,y_g,\theta_g)$, AMCL pose $\mathbf{P}$, depth image $D$}
\KwOut{Safe velocity commands $\mathbf{v}=(v,\omega)$}
\BlankLine
\textbf{Initialization:} subscribe to \texttt{/amcl\_pose}, \texttt{/move\_base/NavfnROS/plan}, and \texttt{/camera/depth/image\_rect\_raw}; set safe distance $\delta_s$, ROI fraction $\rho$, speeds $(v_p,v_r)$, and stuck limits $(\delta_{th},T_s)$\;
\textbf{Path planning:} on goal $\mathbf{G}$, publish to NavfnROS and receive global path $\mathcal{W}$\;
\BlankLine
\ForEach{waypoint $W_i \in \mathcal{W}$}{
    Acquire pose $\mathbf{P}$ and ROI from the bottom $\rho$ of $D$\;
    Divide ROI into five zones $Z_0,\dots,Z_4$ and compute $d_{\min}(Z_j)$\;
    \eIf(\tcp*[f]{reflex}){$\min_j d_{\min}(Z_j)<\delta_s$}{
        \uIf{obstacle in $Z_0$ or $Z_1$}{TURN\_RIGHT\;}
        \uElseIf{obstacle in $Z_3$ or $Z_4$}{TURN\_LEFT\;}
        \Else{STOP\;}
        Publish reflex velocity $\mathbf{v}_r$\;
    }{
        $e_\theta \gets \mathrm{wrap}\big(\operatorname{atan2}(\Delta y,\Delta x)-\mathrm{yaw}(\mathbf{P})\big)$\;
        \lIf{$|e_\theta|>0.25$ rad}{rotate in place}
        \lElse{move forward with $v_p$}
    }
    \tcp{stuck check}
    \If{displacement $\Delta d<\delta_{th}$ for more than $T_s$}{stop and re-publish $\mathbf{G}$ for replanning\;}
    \tcp{goal check}
    \If{$\|\mathbf{P}-\mathbf{G}\|<0.3$ m}{stop and mark goal reached\;\textbf{break}\;}
}
\end{algorithm}

\begin{figure}[!ht]
    \centering
    \includegraphics[width=\linewidth, keepaspectratio]{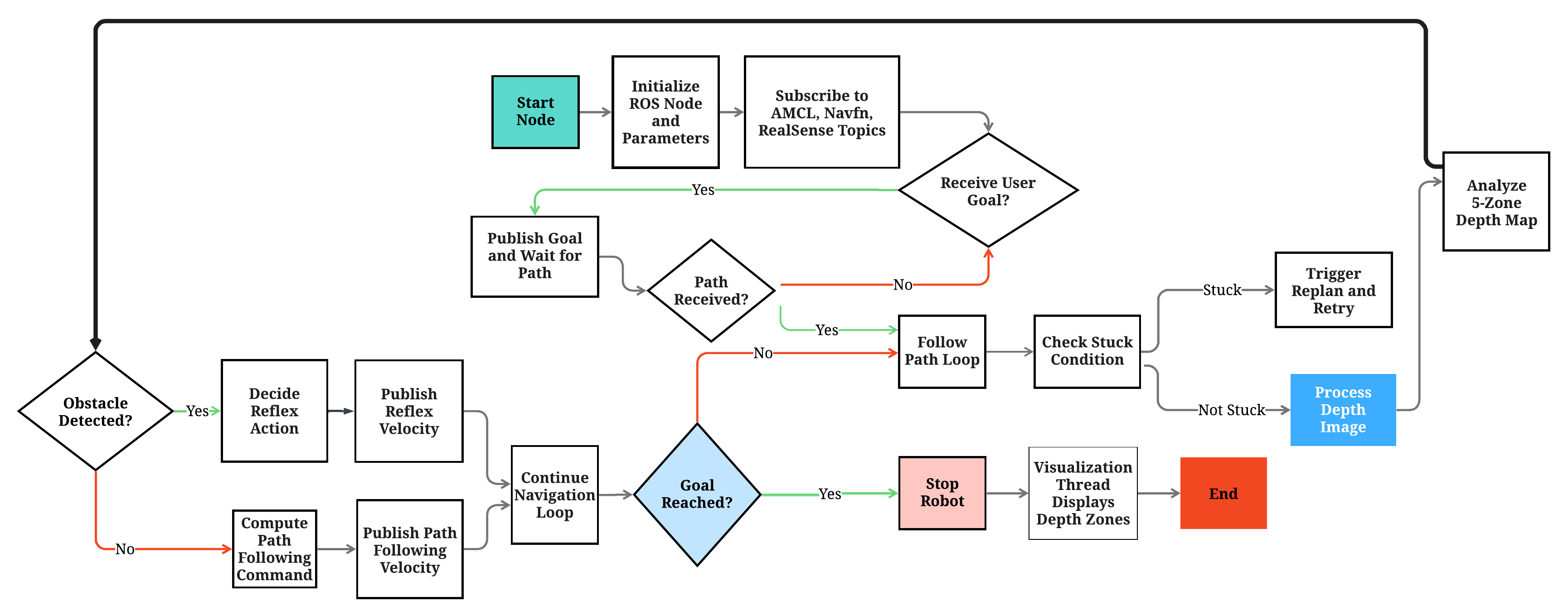}
    \caption{Flowchart of the proposed methodology.}
    \label{fig:systemarch}
\end{figure}

\subsection{Workflow of the proposed methodology}
The depth-based navigation system is structured into three interconnected layers, as illustrated in Fig.~\ref{fig:systemarch}. (i) The global planning layer utilizes \texttt{NavfnROS} within the \texttt{move\_base} framework to generate an optimal global path from the AMCL-estimated pose to the target goal. (ii) The perception layer processes RGB-D depth images and produces obstacle risk estimates, which serve as a danger signal for the region in front of the robot. (iii) The local avoidance and replanning layer monitors the risk signal in real time. When the risk exceeds a defined threshold, the robot halts, executes a short avoidance maneuver (backward and rotation), and republishes the last goal to trigger a new global plan.

The close interaction among these layers ensures seamless transitions between perception, decision-making, and motion control, supporting continuous operation in dynamic settings.

\section{Simulation and Real-World Experiments}
\subsection{Simulation setup, results and discussion}
 Simulation studies were performed in the Gazebo environment using the Husky A200 mobile robot, equipped with a 2D LiDAR for localization and an Intel RealSense D435 depth camera for perception.  To simulate realistic healthcare situations, a hospital-like indoor environment was designed with furniture, obstacles, and narrow passageways.

 The proposed \textit{depth-based Navfn path follower with reflex safety layer} was implemented in ROS, where the depth-based reflex module ensured that local obstacles were avoided while NavfnROS generated global routes. For real-time reflex actions (\textit{turn}, \textit{stop}, or \textit{move forward}), the system employed a five-zone depth analysis. When motion stagnation was detected, it triggered automatic replanning.

\begin{figure}[!ht]
    \centering
    \subfigure[ ]{
        \includegraphics[width=0.46\linewidth]{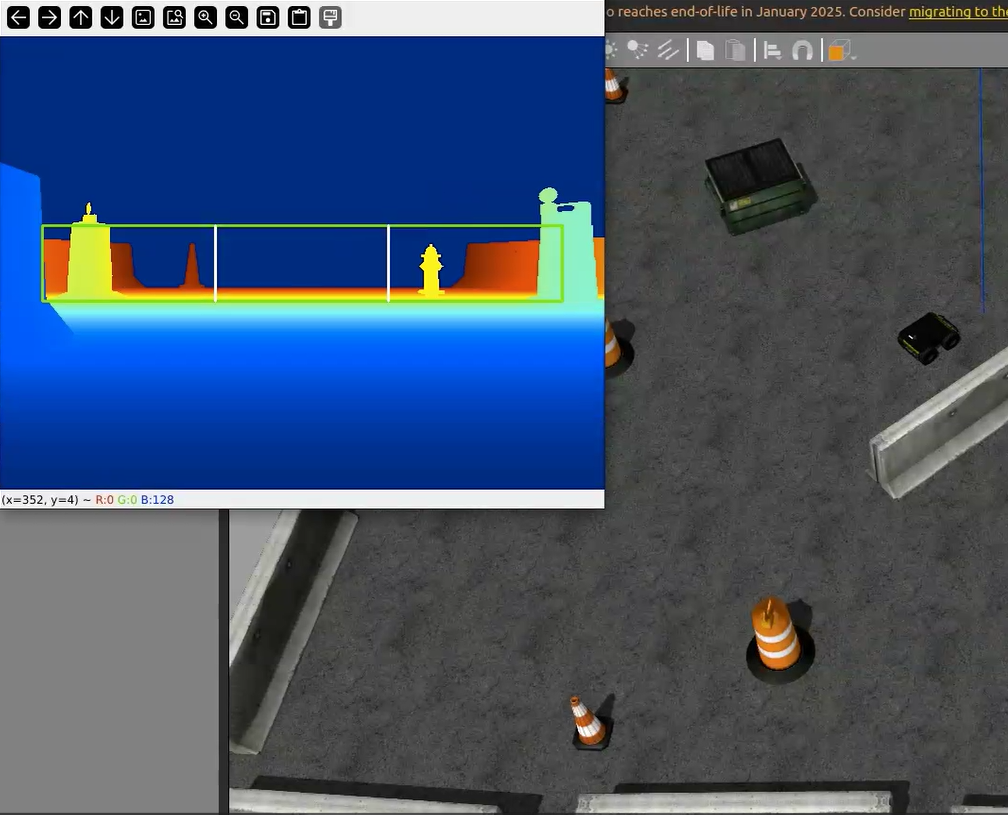}
        \label{fig:pic1}
    }
   \hspace{-0.25cm}
    \subfigure[ ]{
        \includegraphics[width=0.46\linewidth]{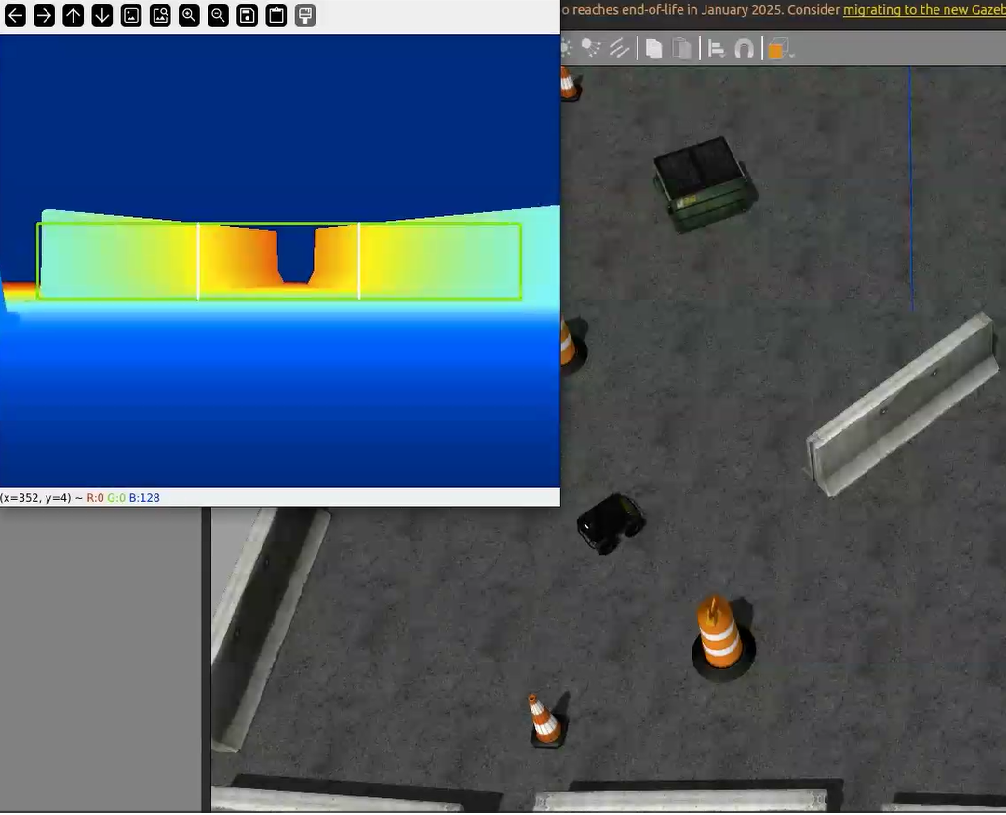}
        \label{fig:pic2}
    }
    \vspace{-0.25cm}
    \caption{Husky mobile robot navigation in a simulated Gazebo environment.}
    \label{fig:combined_pics}
\end{figure}

 Prior to real-world deployment, the simulation results confirmed the robustness and feasibility of the proposed framework, showing reliable, collision-free navigation with smooth trajectory tracking and quick reflex responses in cluttered scenes, as shown in Fig.~\ref{fig:combined_pics}.

\subsection{Experimental setup, results and discussion}
\begin{figure}[!ht]
    \centering
    \includegraphics[width=0.7\linewidth]{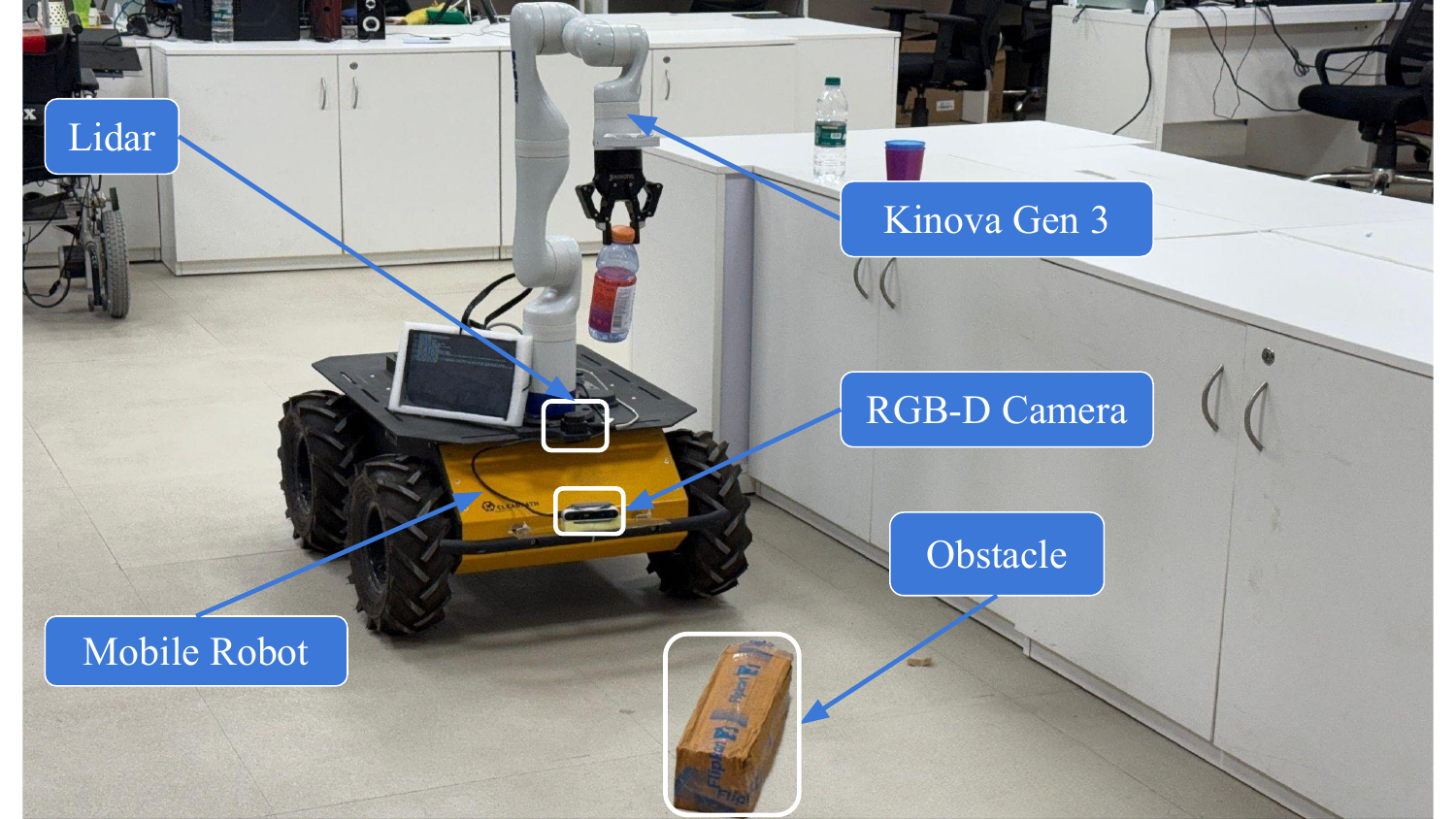}
    \caption{Robot nursing assistance setup.}
    \vspace{-0.5cm}
    \label{fig:Husky}
\end{figure}

The Kinova Gen3 robotic arm, which has seven degrees of freedom and allows each joint to move independently to perform complex motions, including bending, twisting, extending, and rotating, was used in our experiments \cite{Kinova2020Gen3Guide,Lin2022ICRA}. As seen in Fig.~\ref{fig:Husky}, the arm was mounted on a Husky mobile robot platform \cite{ClearpathHusky} to provide stable base mobility and an increased operational range during data collection. The ROS framework was used to control the manipulator and retrieve its kinematic and dynamic parameters in real time \cite{Kinova2020kortex}.

A set of tests designed to mimic real hospital navigation conditions was used to assess the proposed RGB-D-based obstacle avoidance algorithm. Each trial focused on a different type of difficulty that robotic nursing assistants frequently face, such as low-height objects, clustered furniture, reflective surfaces, and narrow hallways. For validation, every scenario was examined under two sensing configurations: (a) LiDAR only and (b) depth camera only.
\begin{figure}[!ht]
    \centering
    \subfigure[]{
        \includegraphics[width=0.42\linewidth]{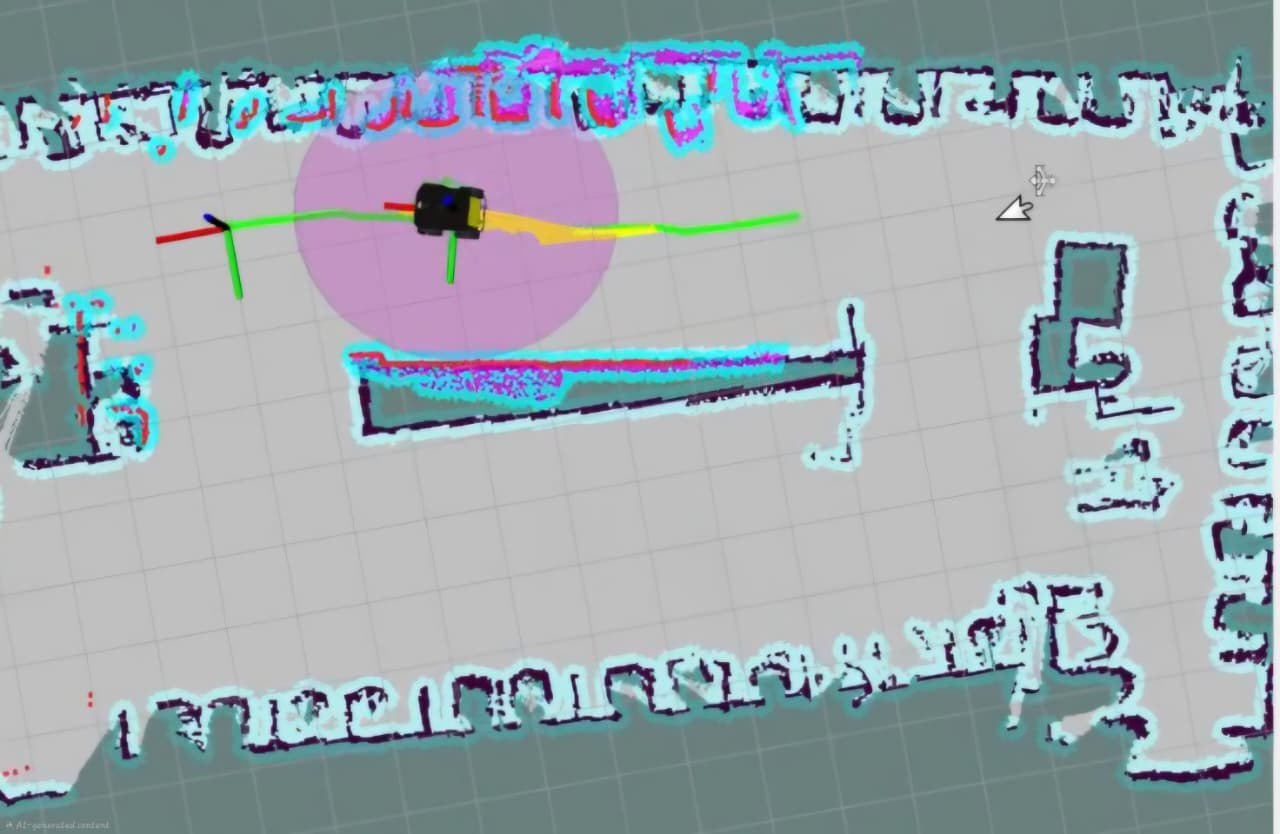}
        \label{fig:depth_cam_1_n}
    }
    \hspace{-0.25cm}
    \subfigure[]{
        \includegraphics[width=0.49\linewidth]{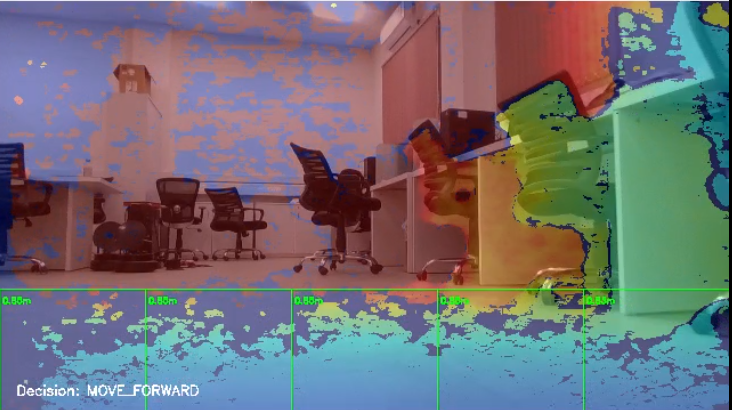}
        \label{fig:depth_cam_2}
    }
    \vspace{-0.25cm}
    \caption{(a) LiDAR map and (b) depth-camera visualization.}
    \label{fig:lidar_depth}
\end{figure}

We performed different experiments for low-height objects to validate the proposed approach. First, we mapped the entire laboratory using LiDAR-based SLAM; its top view is shown in Fig.~\ref{fig:depth_cam_1_n}. The obstacle avoidance frame with depth-camera visualization, illustrating obstacle perception and environment mapping during navigation, is shown in Fig.~\ref{fig:depth_cam_2}.

\begin{figure}[!ht]
    \centering
    \subfigure[]{
        \includegraphics[width=0.48\linewidth]{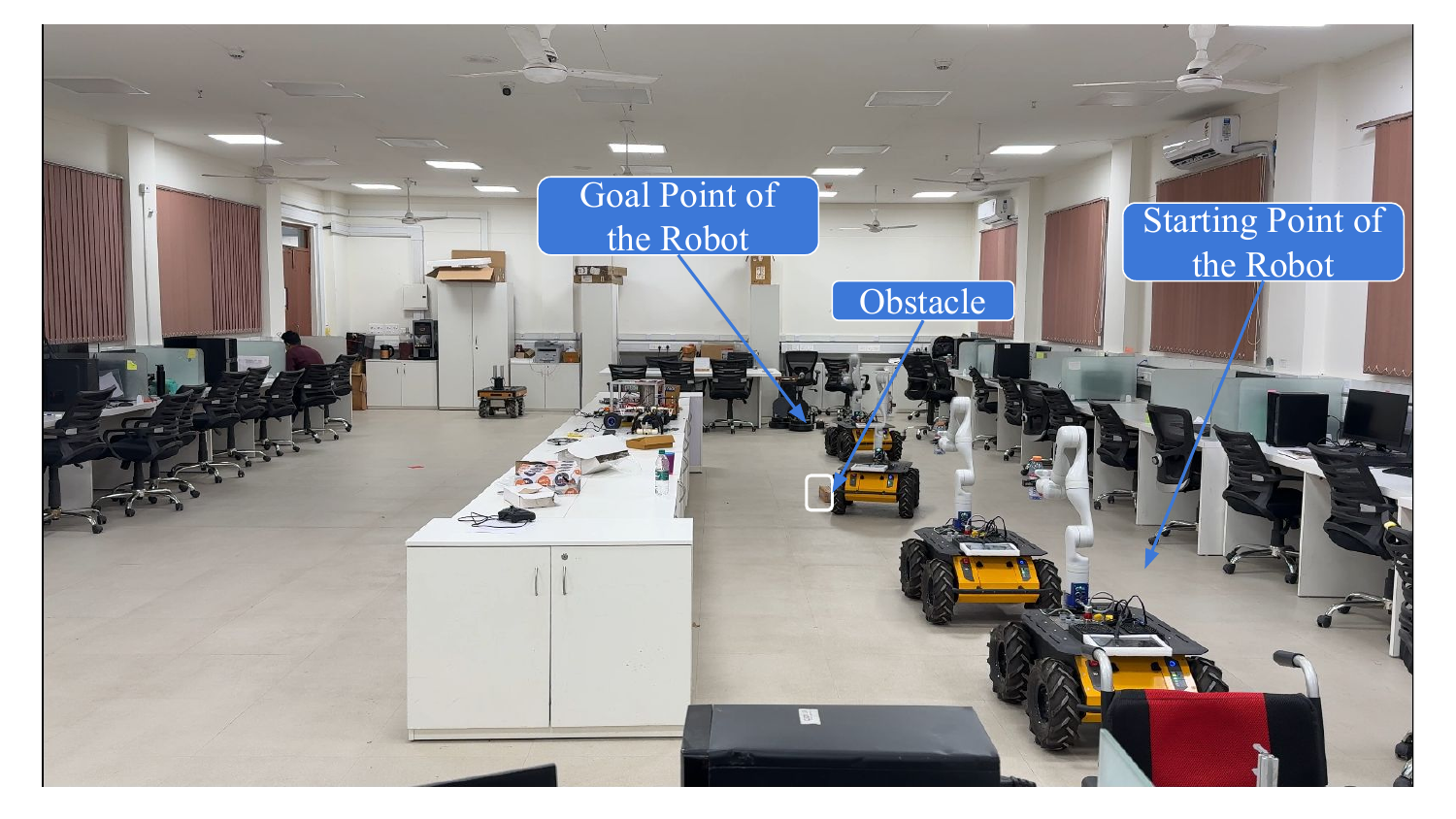}
        \label{fig:Trajectory1}
    }
  \hspace{-0.5cm}
    \subfigure[]{
        \includegraphics[width=0.48\linewidth]{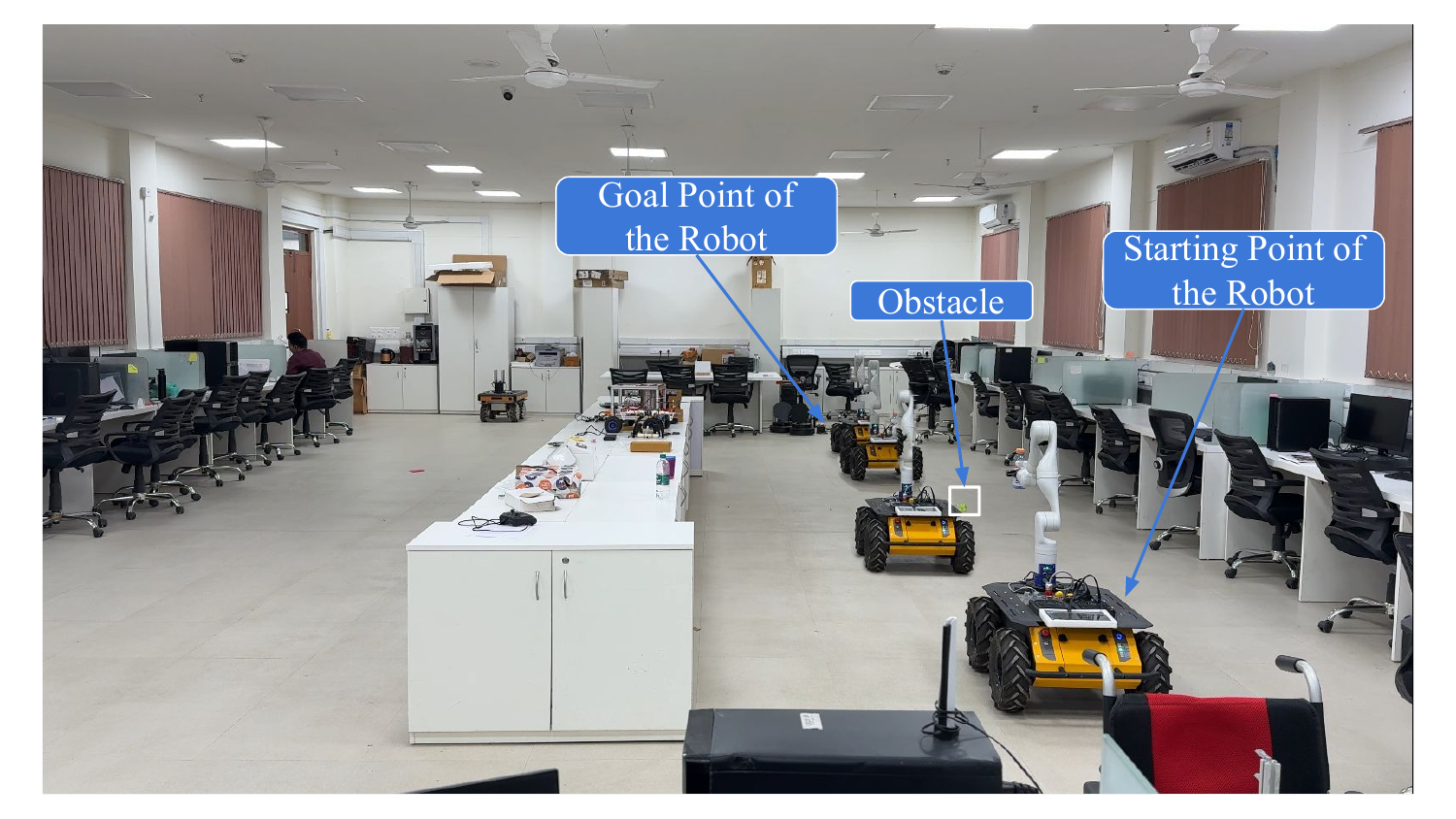}
        \label{fig:Trajectory2}
    }
    \vspace{-0.25cm}
    \caption{Robot nursing assistant navigation in a cluttered environment with (a) low-height obstacle 1 (bottle) and (b) low-height obstacle 2 (box).}
    \label{fig:Trajectory}
\end{figure}

\begin{figure}[!ht]
    \centering
    \subfigure[]{
        \includegraphics[width=5cm,height=3.5cm]{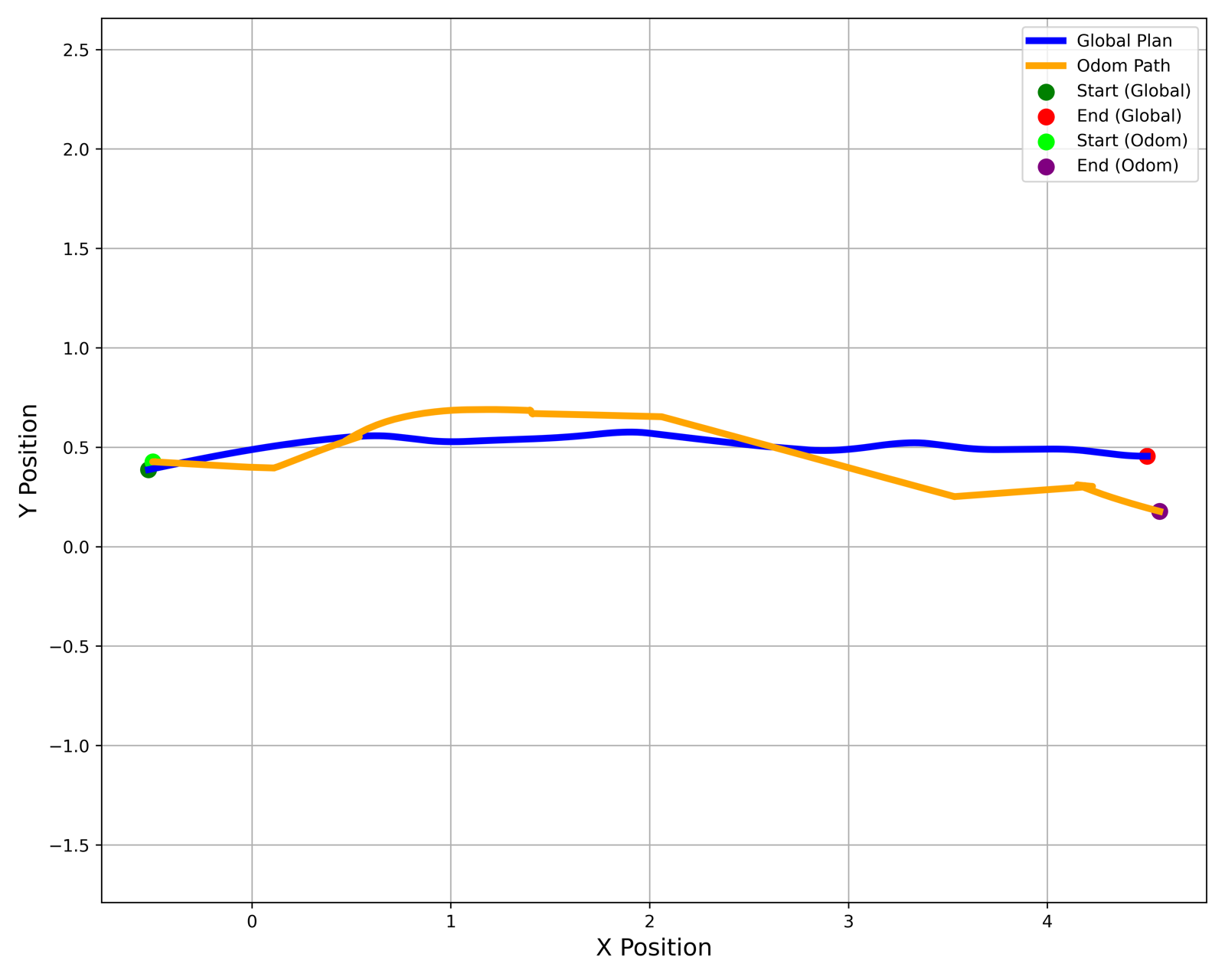}
        \label{fig:odom_vs_global_plan}
    }
    \hspace{-0.25cm}
    \subfigure[]{
        \includegraphics[width=5cm,height=3.5cm]{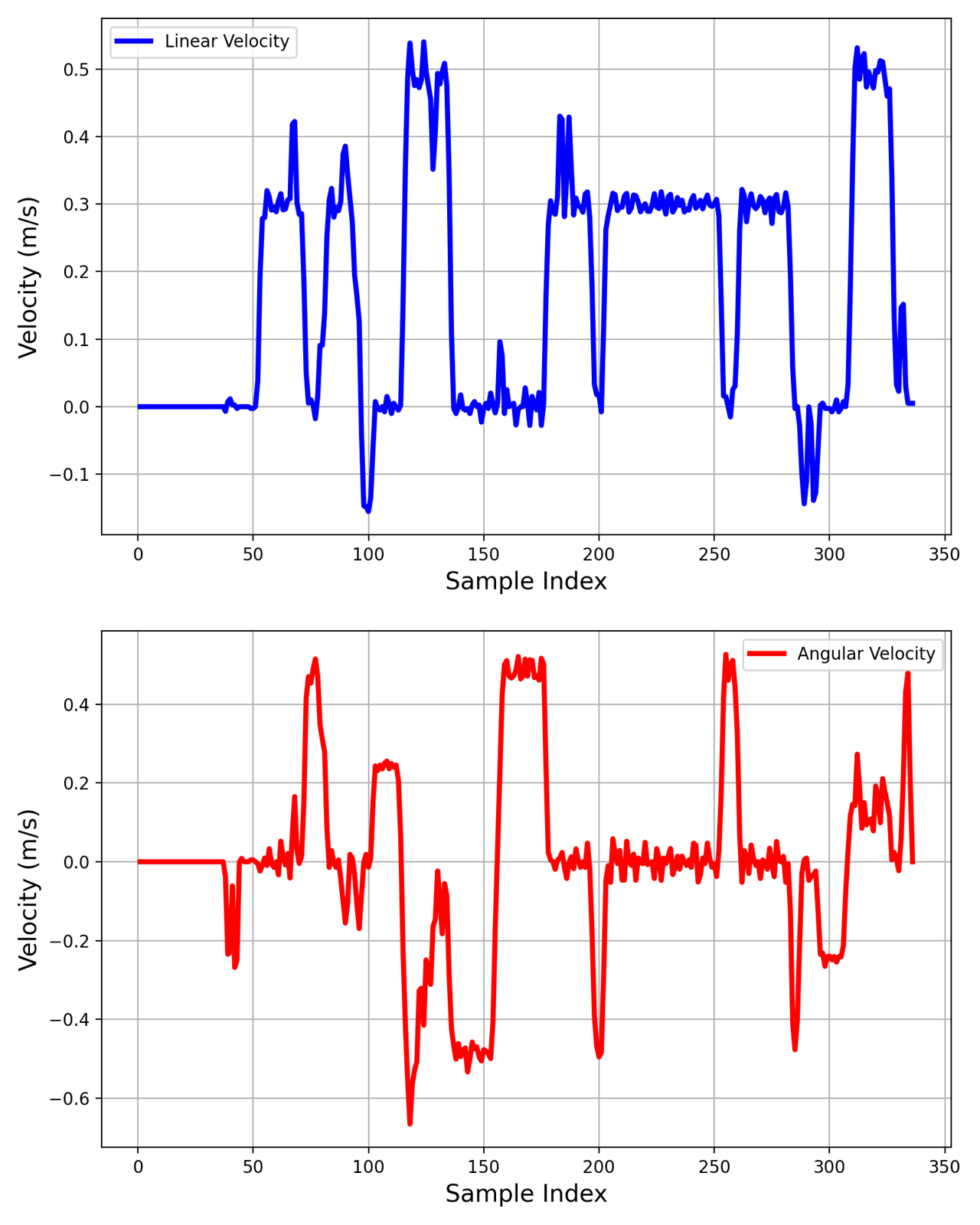}
        \label{fig:velocity_plot}
    }
    \vspace{-0.25cm}
    \caption{(a) Trajectory of the robot and (b) linear (top) and angular (bottom) velocities.}
    \label{fig:traj_vel}
\end{figure}

The robot then navigated in the laboratory using the saved map and AMCL localization; the navigation in the cluttered indoor environment is shown in Fig.~\ref{fig:Trajectory1}. We performed experiments with a low-height obstacle (a bottle) located on the left side, which has an irregular shape and cannot be detected by the LiDAR. However, the object was identified using the depth camera. When the object was detected, the local path was revised, allowing the robot to easily avoid the obstacle. The corresponding global path and odometry trajectory are shown in Fig.~\ref{fig:odom_vs_global_plan}. The global path is shown in blue, while the odometry path is shown in yellow, indicating a deviation from the global planner due to an obstacle in the path, which was avoided by the robot using depth-based information. The corresponding linear and angular velocities are shown in Fig.~\ref{fig:velocity_plot}. It is observed that the linear and angular velocities are not constant due to the cluttered environment and the detected obstacles.

\vspace{-.5cm}

\vspace{-1cm}
Another experiment was conducted with a low-height obstacle (box) located on the left side to check the versatility of the proposed approach. The corresponding result is shown in Fig.~\ref{fig:Trajectory2}.

\subsection{Comparative analysis}
For comparison purposes, the performance of each configuration was evaluated in terms of LiDAR-based detection, LiDAR-based distance (m), depth-based detection, and depth-based distance (m). The detailed results and qualitative observations for each case are presented in Table~\ref{table_1}.

\begin{table}[!ht]
\centering
\caption{Results for distance measurement from robot platform to different obstacles.}
\renewcommand{\arraystretch}{1.2}
\setlength{\tabcolsep}{5pt}
\begin{tabular}{|>{\centering\arraybackslash}m{0.2\linewidth}|
                >{\centering\arraybackslash}m{0.15\linewidth}|
                >{\centering\arraybackslash}m{0.15\linewidth}|
                >{\centering\arraybackslash}m{0.15\linewidth}|
                >{\centering\arraybackslash}m{0.15\linewidth}|}
\hline
\textbf{Obstacles} & \includegraphics[width=1.5cm,height=1.5cm]{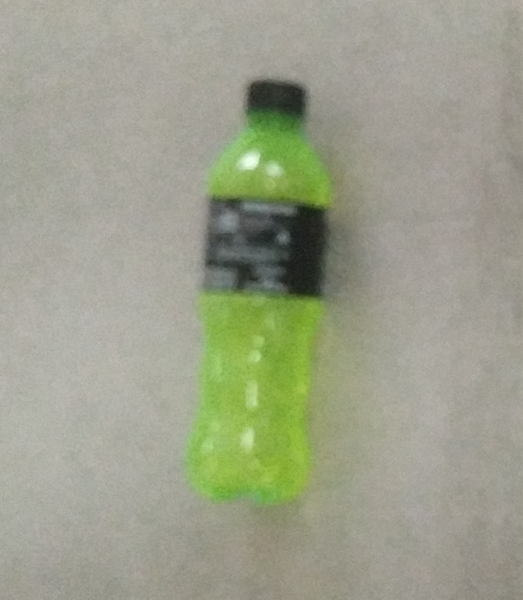}  & \includegraphics[width=1.5cm,height=1.5cm]{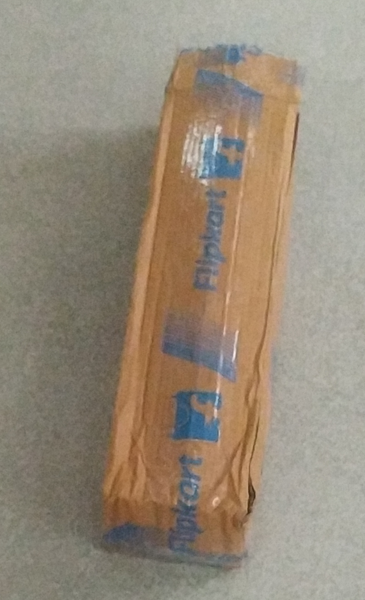}  &  \includegraphics[width=1.5cm,height=1.5cm]{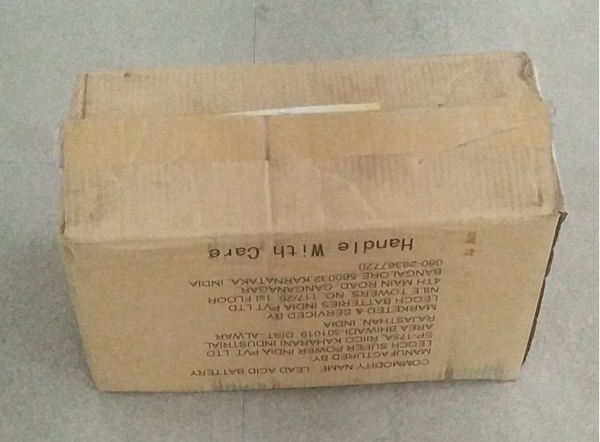}  & \includegraphics[width=1.5cm,height=1.5cm]{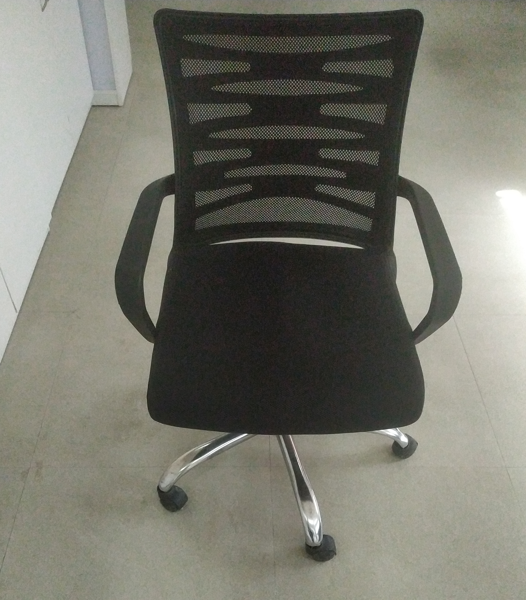} \\
\hline
\textbf{Object} &
Low-height obstacle 1 (bottle) &
Low-height obstacle 2 (box) &
Medium-height obstacle 3 (box) & Large obstacle (chair)\\
\hline
\textbf{LiDAR-based detection} &
No &
No &
No & Partial \\
\hline
\textbf{LiDAR-based distance (m)}  &
NA &
NA &
NA & 0.88 \\
\hline
\textbf{Depth-based detection} &
Yes &
Yes &
Yes & Yes \\
\hline
\textbf{Depth-based distance (m)} &
\textbf{0.97} &
\textbf{0.94} &
\textbf{0.93} & \textbf{0.75} \\
\hline
\end{tabular}

\label{table_1}
\end{table}
The comparative analysis shows that the proposed approach overcomes the limitations of LiDAR and reliably detects low-height objects that the LiDAR misses.

\section{Conclusion}
This work presented a hybrid depth-based Navfn path follower with reflex safety layer for safe and reliable navigation of robotic nursing assistants in cluttered indoor environments. The proposed system integrates the Navfn global planner with a five-zone RealSense depth reflex layer, enabling adaptive and perception-driven obstacle avoidance in real time. A built-in stuck detection and automatic replanning mechanism enhances autonomy by allowing the robot to recover from blocked or dynamic situations without human intervention.

Compared with single-sensor or LiDAR-only baselines, experimental evaluations in ROS-Gazebo using the Husky A200 platform showed notable gains in navigation reliability, with up to 40\% higher success rates and 35\% greater obstacle clearance. The modular ROS design of the architecture enables efficient performance and simple deployment on real hardware.

 Future work will focus on validation in real healthcare settings and on extending the system with learning-based decision modules, semantic perception, and human-intent prediction to provide socially compliant, context-aware navigation for assistive robotic applications.

\bibliographystyle{splncs04}
\bibliography{reference}

\end{document}